\RequirePackage{iftex}
\ifPDFTeX\pdfoutput=1\fi

\documentclass[10pt, logo, copyright]{nvidiatechreport}
\ifXeTeX\microtypesetup{tracking=false}\fi

\usepackage[numbers]{natbib}
\usepackage{soul}
\usepackage{float}
\usepackage{placeins} 
\usepackage{xcolor}
\sethlcolor{yellow}
\hypersetup{allcolors=NvidiaGreen}

\definecolor{nvidiagreen}{HTML}{76B900}

\title{Isekai3D SDG: A Platform for Diverse Synthetic Gameplay Visual Data}
\author{Rafa\l{} Karp, Gene Liu, Shih-Hsin Li, Eric Courtemanche, Shengze Wang, Josef Spjut, David Pankratz, Mischa Kolbe, Michael Stengel,  Sven Elflein, Adam Grygielski, Liwen Wu, Anjul Patney}
\affil{NVIDIA}

\begin{abstract}
\textbf{Abstract:}

We present Isekai3D Synthetic Data Generation (SDG), a platform for
simulating and capturing synthetic 3D gameplay for research in vision and world
models, neural graphics, gameplay agents, and other applications. The platform
combines reusable environment and character assets with Unreal
Engine~\citep{epic_unreal_engine} plugins in a two-stage workflow. First, world
simulation generates gameplay while recording the world state, character
movements, actions, and camera tracks needed for replay. Second, the recorded
simulation is replayed to capture selected outputs, such as rendered video,
images, and depth maps, at a specified resolution and frame rate
under selected rendering and camera configurations. Separating simulation from
capture allows the same recorded gameplay to produce multiple output variants
without repeating the original simulation. Visual outputs are accompanied by
frame-aligned camera, character, and action metadata.
The platform uploads generated data directly in a
training-friendly format, reducing intermediate storage, preparation time,
and manual processing.
This report describes the design of our overall system, including the
platform's assets, plugins, and their functionality, explaining how they
work together to generate diverse synthetic gameplay data.
\end{abstract}

\begin{document}
\maketitle

\section{Introduction}
\label{sec:introduction}
Interactive 3D gameplay provides a setting for studying how visual worlds
evolve and how actions influence them. For vision/world models and gameplay agents,
useful training data can include both rendered observations and the actions,
camera motion, and scene state associated with those observations. A shared
data platform must therefore support variation across environments,
characters, behaviors, and viewpoints while maintaining a consistent
relationship between visual sequences and their accompanying metadata.

Isekai3D SDG addresses this need by providing a platform for building large and diverse synthetic datasets of high-quality 3D gameplay.
The platform combines reusable environment and character assets with Unreal Engine plugins for gameplay behavior,
recording, and capture. This organization allows users to configure
data generation around a common asset collection and workflow, and to
extend individual components as their application requirements change.

The workflow separates recording scene activity from capturing its visual
outputs. A recorded sequence can be replayed with different camera and
rendering configurations, allowing multiple observations to be generated
from the same underlying activity.
This enables diverse applications and use cases, as well as paired
data generation for training purposes.
Capture produces video or images
together with frame-aligned metadata and optional rendering passes such
as depth. Configuration files control asset selection, recording behavior,
and capture variants, supporting automated runs in both editor and
packaged-game workflows.

The workflow builds a downloadable application with selected levels and the
SDG plugins, enabling data generation without installing the Unreal Engine
editor or build tools. Researchers can also implement custom functionality
through multiple extension mechanisms, including their own plugins or
DLL-based extensions.

\begin{figure}[htbp]
  \centering
  \includegraphics[width=\linewidth]{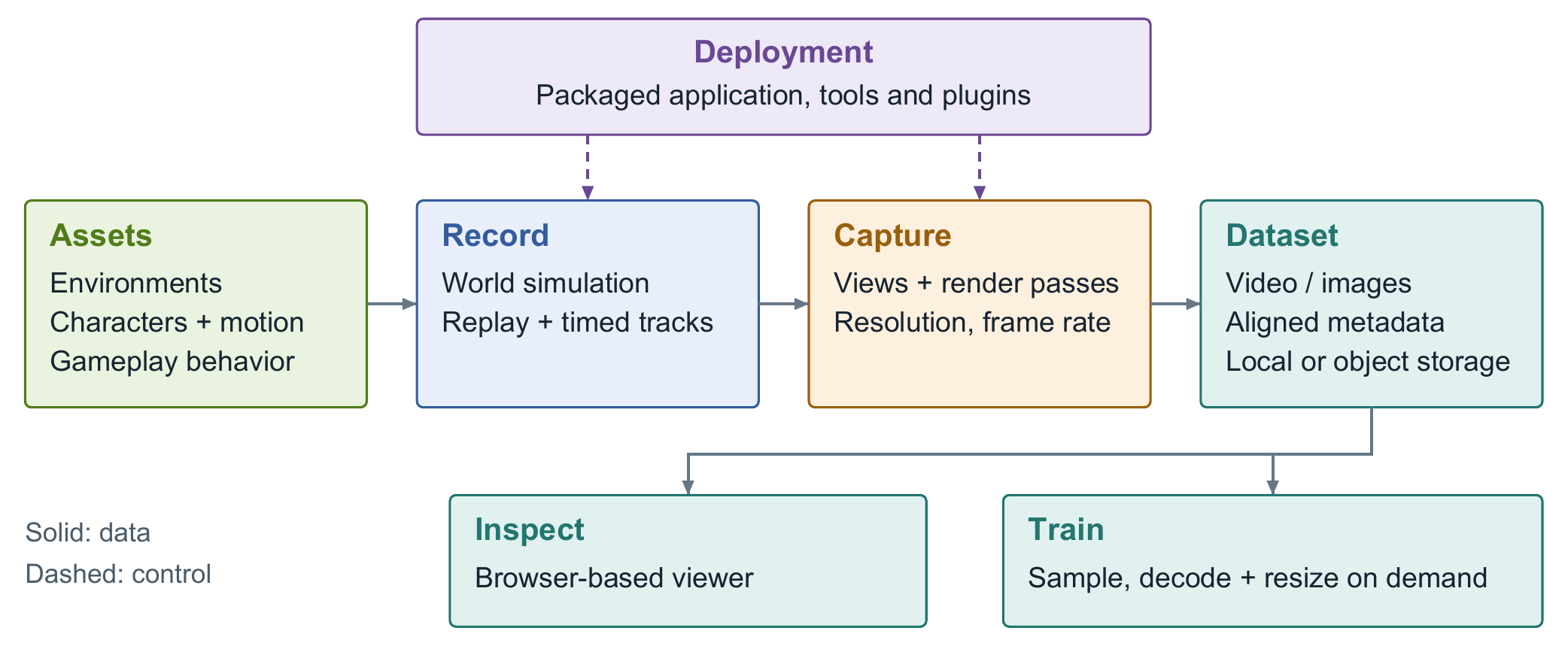}
  \caption{Isekai3D SDG workflow. Recorded gameplay supports
  multiple capture variants, producing aligned visual data and metadata
  for inspection and training.}
  \label{fig:system-overview}
\end{figure}

Figure~\ref{fig:system-overview} summarizes the platform from
asset selection to training.

\section{Approach}
\label{sec:approach}

\subsection{Assets and Environments}

The Isekai3D SDG platform can support varying gameplay and environment categories.
In this report, we describe how we generate third-person  (3P) gameplay
data.
We first describe the two key components of 3P gameplay: characters and environments.

\paragraph{Characters.}
The character collection contains 500+ generated humanoids based on SMPL-X~\citep{pavlakos2019smplx} and Grounded-Exo2Ego~\citep{groundedexo2ego2026}.
We integrate these assets into Unreal Engine through
\textit{IsekaiCharacters}, our custom plugin that packages character
content as a reusable module.
Each character includes a skeletal mesh, materials, and a skeleton
using the same 55-joint SMPL-X layout, with an additional non-deforming
root bone for integration into Unreal Engine. The joint layout covers
the body, hands, jaw, and eyes. A shared animation retargeting setup
transfers motion from Unreal Engine's Manny skeleton to the character
skeletons, allowing the animation library to be reused across the
collection. Characters support a shared library of locomotion and
gameplay animations for varied movement and actions.

\begin{figure}[tbp]
    \centering
    \includegraphics[
        width=0.95\linewidth,
        height=0.32\textheight,
        keepaspectratio
    ]{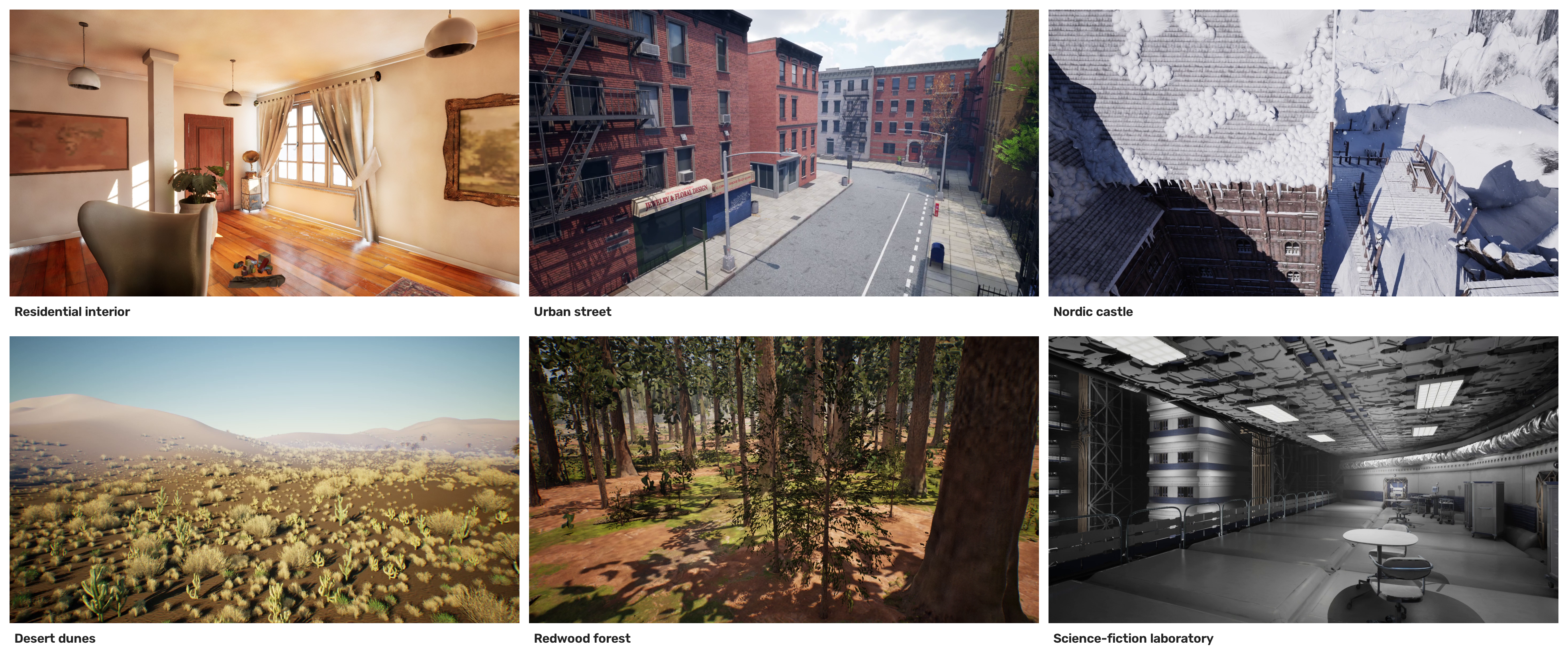}
    \caption{Representative Isekai3D environments. Top row:
    residential interior, urban street, and Nordic castle.
    Bottom row: desert dunes, redwood forest, and science-fiction
    laboratory.}
    \label{fig:environment_diversity}
\end{figure}

\paragraph{Environments.}
Isekai3D provides 245+ environments sourced from Fab~\cite{epic_fab}
and acquired under the applicable asset licenses. These environments
span residential interiors, urban streets, natural landscapes, and
stylized settings, providing variation in scene geometry, materials,
lighting, and background clutter. This diversity supports synthetic
data generation across different visual contexts and evaluation of
model robustness to changes in the surrounding environment.
Figure~\ref{fig:environment_diversity} shows six representative examples.

\subsection{World Simulation and Recording}
World simulation generates gameplay in a selected environment using
configured characters and behaviors. Our custom \textit{IsekaiSceneReplay} plugin records the simulation using Unreal Engine's replay system, 
supplemented with timestamped player actions and camera trajectories, 
while \textit{IsekaiGameplay} supplies configurable gameplay actions and interactions. 
Automated recording uses a fixed simulation timestep and can run without rendering through Unreal
Engine's NullRHI mode, avoiding image-generation work until capture.
The recording stage produces an Unreal replay file containing replicated
world state and a companion JSON file containing timestamped character
positions, orientations, and velocities; camera poses and fields of view;
and simulated input events, including action presses, releases, and axis
values. These files retain the simulated activity for subsequent capture.

An optional \textit{phantom simulation} mode improves data generation performance by recording multiple independent character trajectories in a shared world simulation. Characters collide with the environment but do not collide with or otherwise affect one another; each capture renders only the selected character. 
Sharing the recording stage across trajectories reduces repeated simulation work, with observed end-to-end throughput improvements of approximately $1.3$--$2\times$, depending on the machine and workload. The benefit is greater when recording accounts for a substantial fraction of generation time; per-character behavior, animation, and capture still incur computation.

\subsection{Capture and Data Outputs}
Capture uses the saved replay and timestamped tracks rather than rerunning
the autonomous behavior decisions. At each capture timestep, the system
restores the recorded character trajectory, samples the selected camera
track, and reconstructs the corresponding input and action state. This
provides deterministic sampling of the recorded camera, character, and
input timelines across capture runs using the same settings. The same
recording can therefore support different output resolutions and rendering
passes, while camera variants can be selected explicitly without repeating
the original world simulation.

\begin{figure}[tbp]
  \centering
  \IfFileExists{assets/capture_outputs_grid.png}{%
    \includegraphics[width=0.90\linewidth]{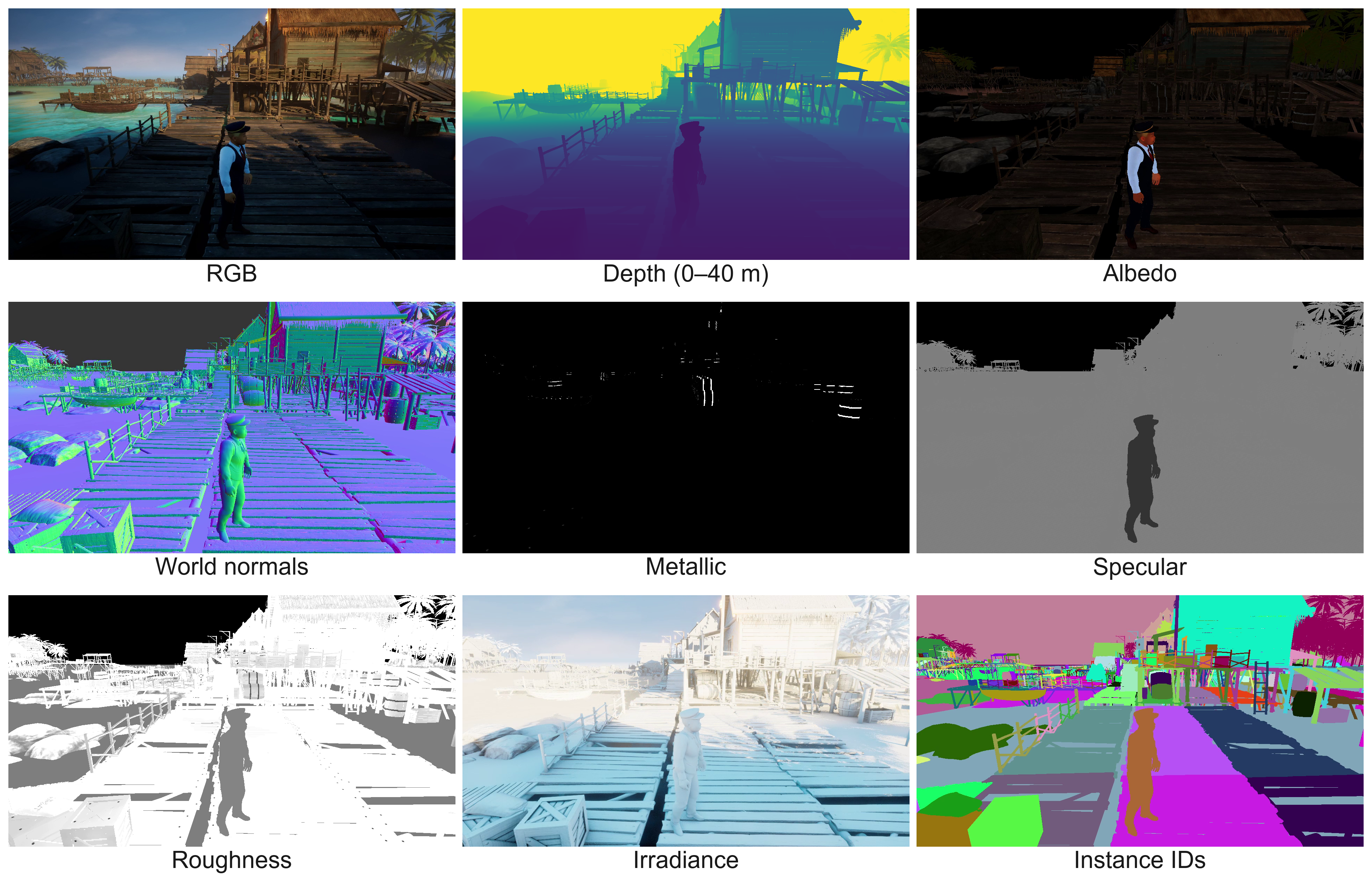}%
  }{%
    \fbox{\parbox[c][0.58\linewidth][c]{0.94\linewidth}{\centering
      Capture output grid\\[1em]
      Image file: \texttt{assets/capture\_outputs\_grid.png}\\[1em]
      The image is displayed when the file is available to the compiler.}}%
  }
  \caption{Capture outputs for a single gameplay frame. From left to right
  and top to bottom: RGB, depth, albedo, world-space normals, metallic,
  specular, roughness, irradiance, and instance IDs. Depth is color-mapped
  from 0 to 40\,m, with purple indicating nearer surfaces and yellow
  indicating distances of 40\,m or greater.}
  \label{fig:capture-outputs}
\end{figure}

Figure~\ref{fig:capture-outputs} illustrates nine capture outputs from the
same recorded frame and camera viewpoint, including RGB, depth, material
properties, world-space normals, irradiance, and instance IDs.

\begin{figure}[htbp]
  \centering
  \includegraphics[width=\linewidth]{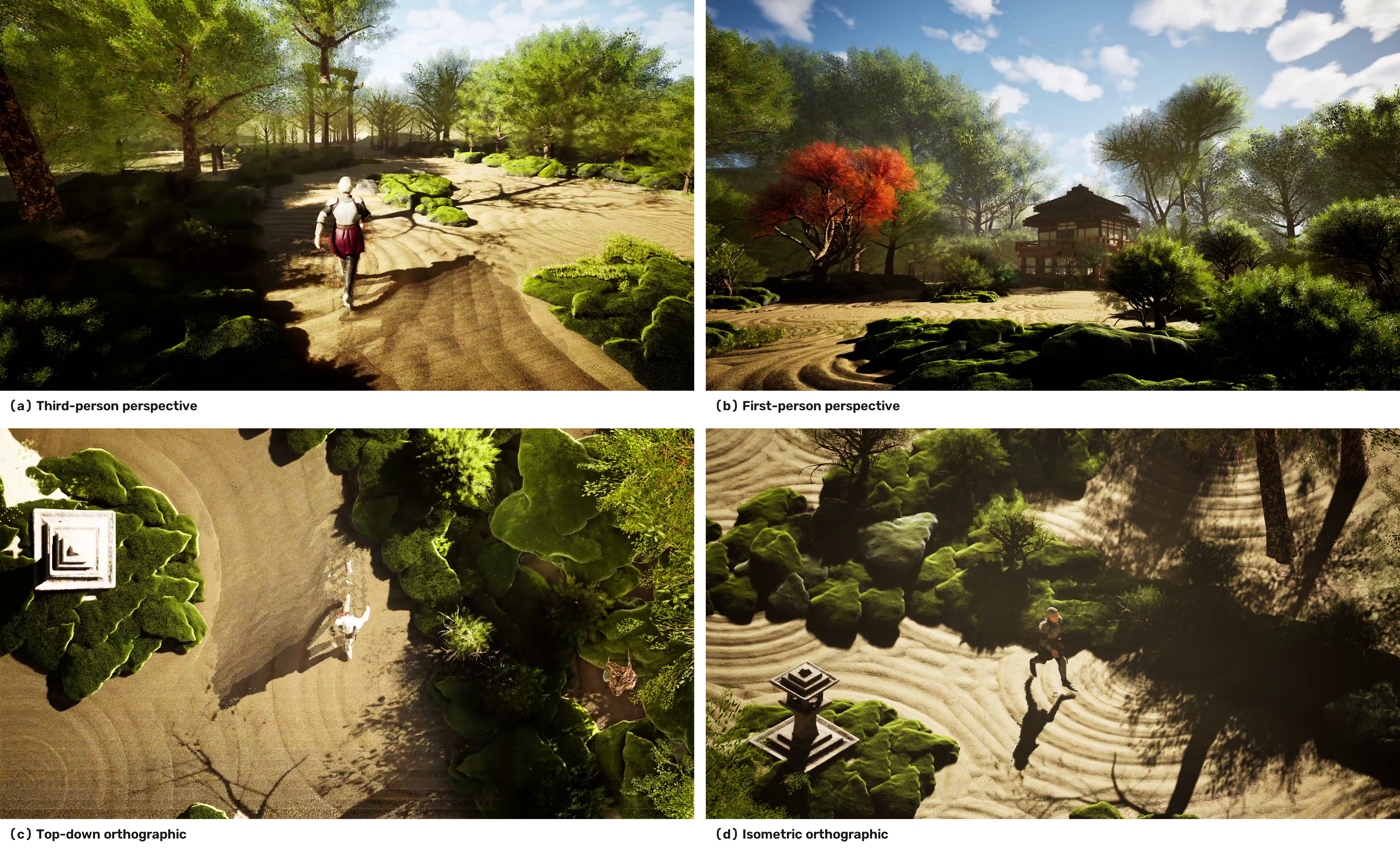}
  \caption{Camera configurations applied to the same recorded
  gameplay moment. Top row: third-person and first-person perspective
  views. Bottom row: top-down and isometric orthographic views.}
  \label{fig:camera-modes}
\end{figure}

\begin{figure}[htbp]
  \centering
  \includegraphics[width=\linewidth]{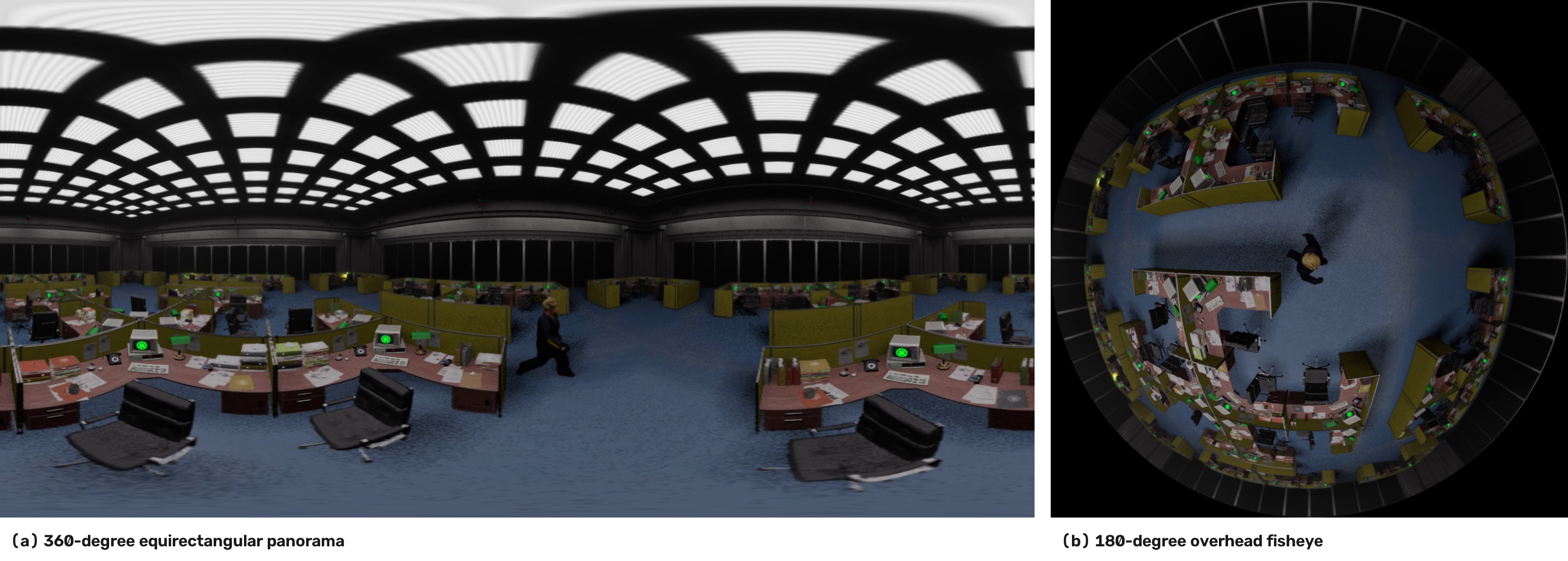}
  \caption{Projection modes applied to the same recorded gameplay
  moment in an office environment. Left: a 360-degree equirectangular
  panorama, horizontally re-centered for display. Right: a
  180-degree fisheye view from above the character.}
  \label{fig:projection-modes}
\end{figure}

Camera and projection settings allow the same recorded gameplay to
provide complementary observations without repeating the world
simulation. Third-person and first-person perspective views capture
character-level activity and the scene from the character's viewpoint,
while top-down and isometric orthographic views emphasize spatial
layout and movement through the environment
(Figure~\ref{fig:camera-modes}). Additional projection modes include
360-degree equirectangular panoramas for capturing the surrounding
environment and 180-degree overhead fisheye views for wide-angle
coverage around the character (Figure~\ref{fig:projection-modes}).
Together, these options support datasets with varied viewpoints and
image geometries, including paired observations of the same recorded
activity with frame-aligned camera, character, and action metadata.

Standard captures use hardware-encoded H.264 video with variable
bitrate (15\,Mb/s target, 18\,Mb/s configured maximum), packaged as MP4
without re-encoding. Path-traced captures use H.264 at CRF~6 with 8-bit
4:2:0 pixels (\texttt{yuv420p}). For perspective and orthographic views,
supported radiance passes can instead be stored losslessly as scene-linear
EXR frames, bypassing tonemapping and video compression.
Standard browsers and video libraries can inspect the MP4 files directly. A thin file-I/O adapter maps reads
to object-store range requests, allowing training to read selected clips
without first copying the full dataset locally.

\subsection{Dataset Format and Training}

Preparing fixed training clips in advance incurs transcoding
and storage costs and ties the dataset to chosen resolutions, frame rates,
and clip lengths. Changing these choices can require another preparation
pass. Custom file formats can also make samples harder to inspect.

Isekai3D supports lazy datasets over stored videos and metadata.
Researchers select resolution, frame rate, clip duration, and filtering
criteria when launching training; clips are read and transformed on demand.
In our measured workflow, constructing a lazy dataset from 10,000 hours of
video took 3 minutes using Python. Copying the same amount of footage without
processing took 16 hours with a tuned copying utility. Reading and
transforming samples still incur cost during training.

In another example, users were able to work with more than 30 lazy
datasets, often combining several in a training run. If each required a full copy at
the measured rate, 30 copies would take 480 aggregate hours, compared with
1.5 hours for 30 three-minute dataset constructions. This comparison
estimates preparation time under the full-copy assumption. Reusing the
stored videos also avoided hundreds of terabytes of duplicate storage.

\subsection{Deployment and Extensibility}
The full Unreal Engine development setup, including the project, assets,
and supporting tools, requires approximately 1\,TB of disk space and
substantial setup effort. To make data generation accessible without this
development environment, we provide a packaged game that includes our
Unreal Engine plugins and assets, providing the full SDG functionality.
Levels are packaged separately, allowing researchers to
download the base application and only the environments needed for their
experiments. Automated recording and capture run directly against this
packaged game, without requiring an Unreal Engine installation or local
compilation. Runtime configuration and automation interfaces allow researchers to programmatically steer data generation and distribute varied experiments across multiple machines using the same packaged build.

\begin{figure}[H]
  \centering
  \includegraphics[width=\linewidth]{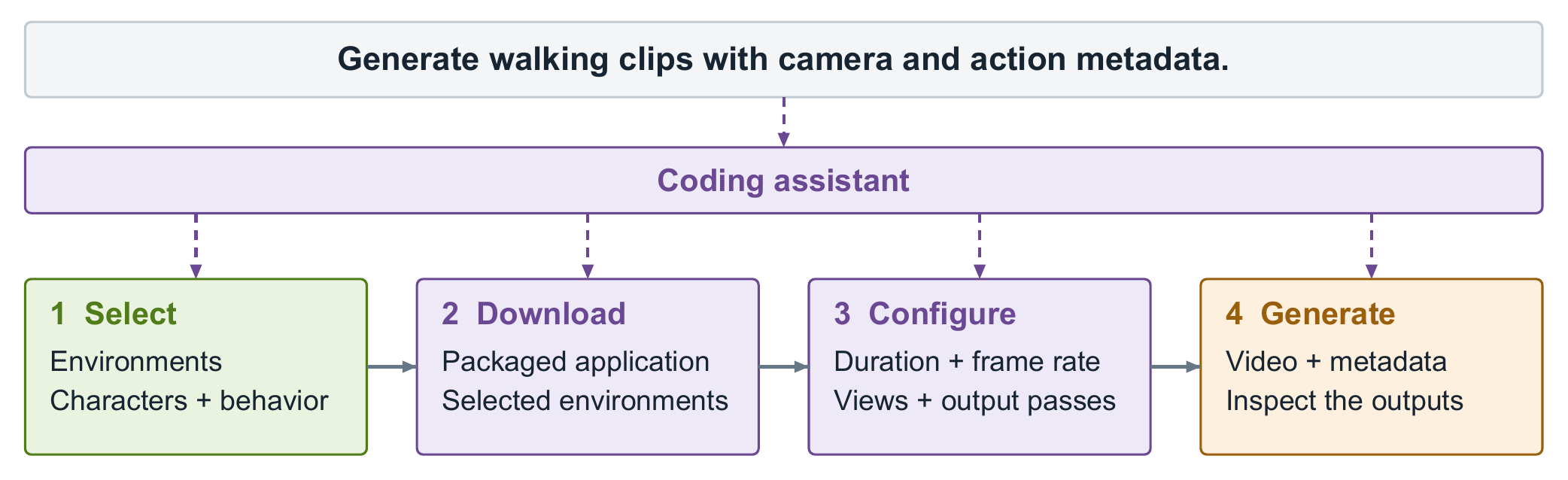}
  \caption{Example workflow: a coding assistant selects content,
  configures generation, and checks the outputs.}
  \label{fig:agent-generation}
\end{figure}

\Needspace{6\baselineskip}
\paragraph{Agent-assisted generation.}
A coding assistant can use the command-line tools and
configuration files to select environments and characters, download the
required content, set recording and capture options, and run generation
(Figure~\ref{fig:agent-generation}). It can then inspect the outputs and
adjust the configuration. SPEAR~\citep{spear2026} also demonstrates
coding-assistant use with programmable simulators.

The game uses a modular build in which components are distributed as
dynamic-link libraries (DLLs). Researchers can use a smaller Unreal Engine
project to compile modified versions of these components with compatible
engine and build tools, then use the resulting DLLs with the packaged
application. This supports custom functionality without requiring the
full environment collection and development project for each change.

\begin{figure}[htbp]
    \centering
    \includegraphics[
        width=0.78\linewidth,
        height=0.28\textheight,
        keepaspectratio
    ]{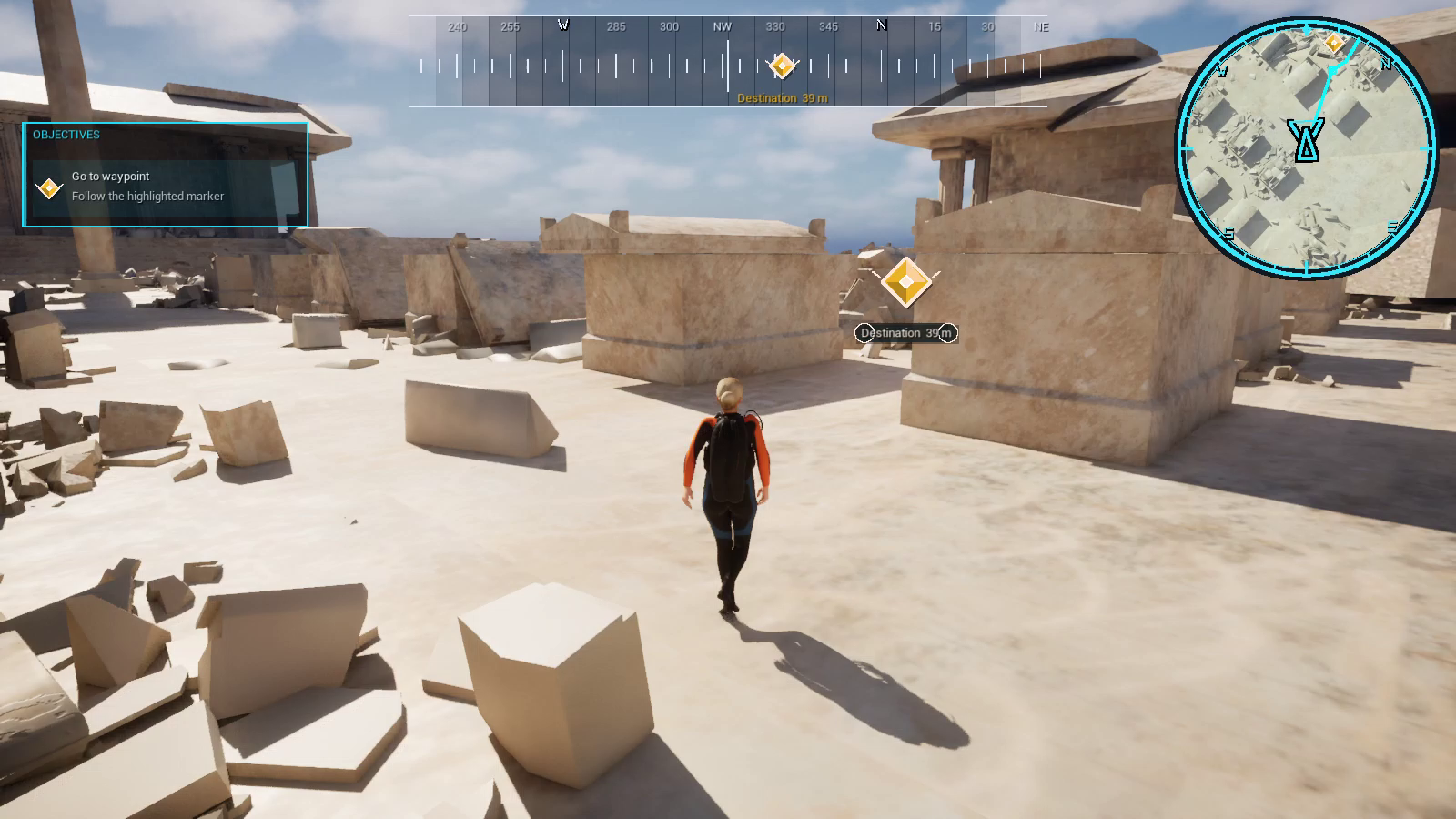}
    \caption{Navigation UI rendered by the external IsekaiGFMUI
    plugin, including a waypoint marker, compass, minimap with
    route guidance, and an example objective panel.}
    \label{fig:navigation-ui}
\end{figure}

We also provide an automated integration workflow for custom Unreal Engine
plugins. Researchers can add their own plugins to extend the SDG system
or reuse the platform's assets in their own workflows. For example, the
external \textit{IsekaiGFMUI} plugin adds configurable waypoint markers,
compass indicators, minimaps, and objective panels to captured gameplay.
Figure~\ref{fig:navigation-ui} illustrates these elements together,
showing how an external plugin can enrich rendered observations with
navigation and task information. Together, these deployment and extension
options support both users who need a ready-to-run data generator and
developers who want to adapt individual components independently.

\FloatBarrier
\Needspace{13\baselineskip}
\section{Conclusion}
\label{sec:conclusion}
Isekai3D SDG presents a modular and extensible platform for synthetic data
generation of game and game-like scenes.
It builds on reusable assets and Unreal Engine
plugins in a shared environment for achieving flexibility and scale.
Separating world simulation and recording from capture allows
the same gameplay to be reused across camera configurations, resolutions,
and output types while retaining aligned metadata. The packaged application
makes data generation accessible without installing the Unreal Engine
editor, while custom plugins or DLL-based extensions allow researchers to
adapt functionality independently. Together, these
capabilities provide a foundation for creating datasets tailored to research
in world models, gameplay agents, and other applications.

\section{Acknowledgements}
We thank Christian Laforte for his support of the synthetic
data effort.


\end{document}